\documentclass{article}
\usepackage{spconf,amsmath, amsfonts, epsfig, float}
\usepackage{url, subcaption}
\usepackage[capitalise]{cleveref}
\usepackage{algorithm}
\usepackage{algpseudocode}

\title{NEURAL EMBEDDING COMPRESSION FOR EFFICIENT MULTI-TASK EARTH OBSERVATION MODELLING}
\twoauthors
{Carlos Gomes
\thanks{
  Co-funded by the European Union (Horizon Europe, Embed2Scale, 101131841).
  \\ © 2024 IEEE. Personal use of this material is permitted. Permission from IEEE must be obtained for all other uses, in any current or future media, including reprinting/republishing this material for advertising or promotional purposes, creating new collective works, for resale or redistribution to servers or lists, or reuse of any copyrighted component of this work in other works.
}
}
  {cpi@zurich.ibm.com\\IBM Research -- Europe\\ Switzerland}
  {Thomas Brunschwiler}
  {tbr@zurich.ibm.com\\IBM Research -- Europe\\ Switzerland}
\begin{document}
%
\maketitle

\begin{abstract}

As repositories of large scale data in earth observation (EO) have grown, so have transfer and storage costs for model training and inference, expending significant resources.
We introduce \textit{Neural Embedding Compression (NEC)}, based on the transfer of compressed embeddings to data consumers instead of raw EO data.
We adapt foundation models (FM) through learned neural compression to generate multi-task embeddings while navigating the tradeoff between compression rate and embedding utility.
We update only a small fraction of the FM parameters ($\sim$$10\%$) for a short training period ($\sim$$1\%$ of the iterations of pre-training).
We evaluate \textit{NEC} on two EO tasks: scene classification and semantic segmentation.
Compared with applying traditional compression to the raw data, \textit{NEC} achieves similar accuracy with a $75\%$ to $90\%$ reduction in data.
Even at $99.7\%$ compression, performance drops by only $5\%$ on the scene classification task.
Overall, \textit{NEC} is a data-efficient yet performant approach for multi-task EO modelling.
\end{abstract}

\section{INTRODUCTION}
\label{sec:intro}
\begin{figure}[t]
\centering
\includegraphics[width=0.8\linewidth]{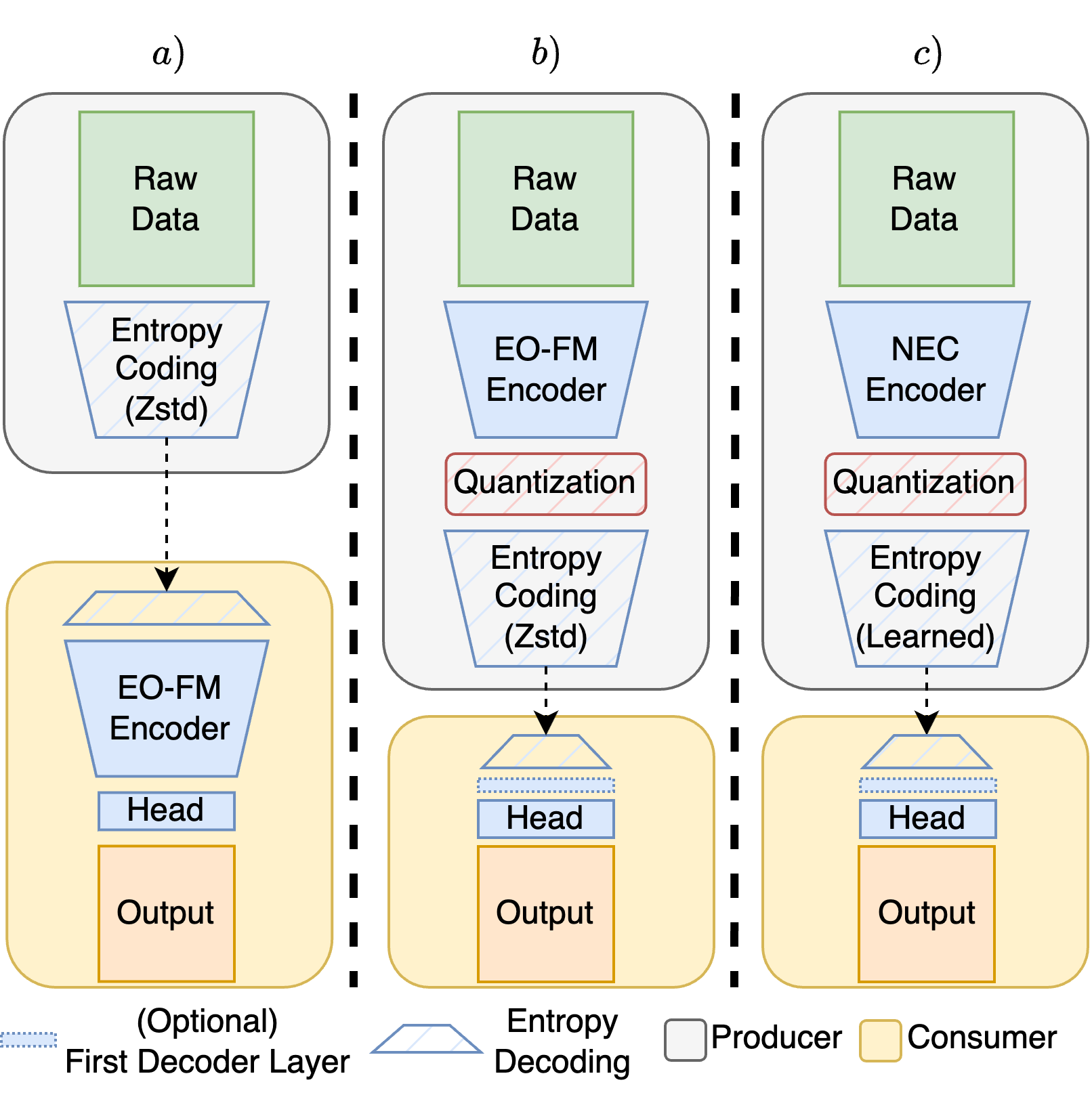}
  \caption{Downstream comparison: a) \textit{Raw Data Compression~(RDC)};  b) \textit{Uniformly Quantized Embeddings~(UQE)}; c) \textit{Neural Embedding Compression~(NEC)}. `Learned' refers to entropy coding with the distribution learned during training.}
    \label{fig:downstream_finetuning}
\end{figure}

Earth observation (EO) repositories comprise some of the largest data stores globally. 
Due to their widespread use, these repositories experience extremely large volumes of data transfers.
For example, users of the Sentinel Data Access System downloaded 78.6 PiB of data in 2022 alone~\cite{sentinel_access}.
The transfer of such data volumes between data producers and consumers (\cref{fig:downstream_finetuning}~(a)) causes substantial latency and requires significant amounts of energy and vast storage capacities. 

Due to its scale, manual processing of the incoming EO data is not an option, leading to the adoption of machine learning methods, including deep learning.
Recently, several EO foundation models (EO-FM) based on self-supervised learning (SSL), pre-trained on EO data sampled across the planet, have been introduced~\cite{ssl4eo, rvsa, ringmo, prithvi, satmae, scalemae}. 
The general-purpose embeddings they produce have been demonstrated as suitable for training task-specific heads for multiple downstream tasks (e.g. scene classification, semantic segmentation). 
The approach requires fewer annotated samples to train the task-specific model and achieves superior performance and generalization to unseen data.
On the other hand, FMs consist of hundreds of millions or even billions of parameters~\cite{prithvi, cha2023billion} and thus are computationally demanding during pre-training and at inference time.

To overcome these issues, we propose the exchange of EO-FM embeddings computed at the data provider instead of the raw data. 
It follows that novel data only has to be processed once by the computationally expensive encoder.
The resulting embeddings can be stored and reused for multiple downstream applications. 
The inference can be performed at the data consumer with a lightweight head.

One of the main challenges is now the efficient exchange of embeddings, which we address in this work.
This is relevant as FMs do not have compression as a goal, often producing embeddings that are even larger than the original data.
When transferring EO data, lossless compression is usually applied, as in~\cref{fig:downstream_finetuning}~(a), resulting in a compressed representation of the data from which it can be reconstructed with no information loss.
We refer to this method as \textit{Raw Data Compression (RDC)}.
%

One could consider applying lossless compression to the embeddings as well.
For higher compression rates, lossy compression methods could be employed at the cost of some distortion.
A suitable metric for this distortion is critical in elaborating a lossy compression method.
To minimize the distortion, different codecs (e.g., JPEG~\cite{jpeg} for images, MP3~\cite{mp3} for audio) exist to compress data in different domains.
Each of these minimizes a domain-specific metric for distortion, usually some proxy for human perception.
Currently, no such codec exists in the domain of embeddings for downstream tasks.

As a simple approach, depicted in~\cref{fig:downstream_finetuning}~(b), we could consider quantizing the embeddings to different bit levels before entropy coding.
We refer to this as \textit{Uniformly Quantized Embeddings (UQE)}.

In this study, we propose the application of learned neural compression, which has been well studied in the domain of image and video compression~\cite{balle2016end, balle2018variational, vct, lu2018dvc}. 
This data-driven approach allows us to learn an encoder that produces compressible multi-task embeddings from data while minimizing an appropriate distortion metric: the loss across downstream tasks.
We call this approach \textit{Neural Embedding Compression (NEC)} and illustrate it in~\cref{fig:downstream_finetuning}~(c).

Our contributions in this paper are threefold: \textbf{i)} The discussion on EO-FM embedding sharing, \textbf{ii)} the integration of neural compression into the data pipeline for classification and segmentation tasks, and \textbf{iii)} the benchmarking \textit{NEC} approach on two downstream EO applications.

\section{RELATED WORK}
\label{sec:Related Work}
Learned neural compression, popularized by Ball\'{e}~\emph{et al.}~\cite{balle2016end}, aims to integrate information theoretical concepts into neural network training for data compression.
A variety of works followed proposing more powerful entropy models~\cite{balle2018variational, joint_balle} and extending its applications to video compression~\cite{lu2018dvc,scale_space, vct} and even neural network compression~\cite{oktay_scalable_2020}.
%

Singh~\emph{et al.}~\cite{singh2020end} explore learned compression for embeddings, but produce task-specific encoders. Dubois~\emph{et al.}~\cite{dubois2021lossy} lay out the theoretical foundation for learned compression of predictive features and adapt CLIP~\cite{CLIP} to produce compressed embeddings for generic classification tasks. 
We expand this scope to explore compressed embeddings for pixel-wise prediction as well as classification. We target EO applications as input data is often sourced from queries on large remote repositories, rather than from local user input, making it a good fit for \textit{NEC}.

\section{METHOD}
\label{sec:method}

\begin{figure}
\centering
{\includegraphics[width=\linewidth]{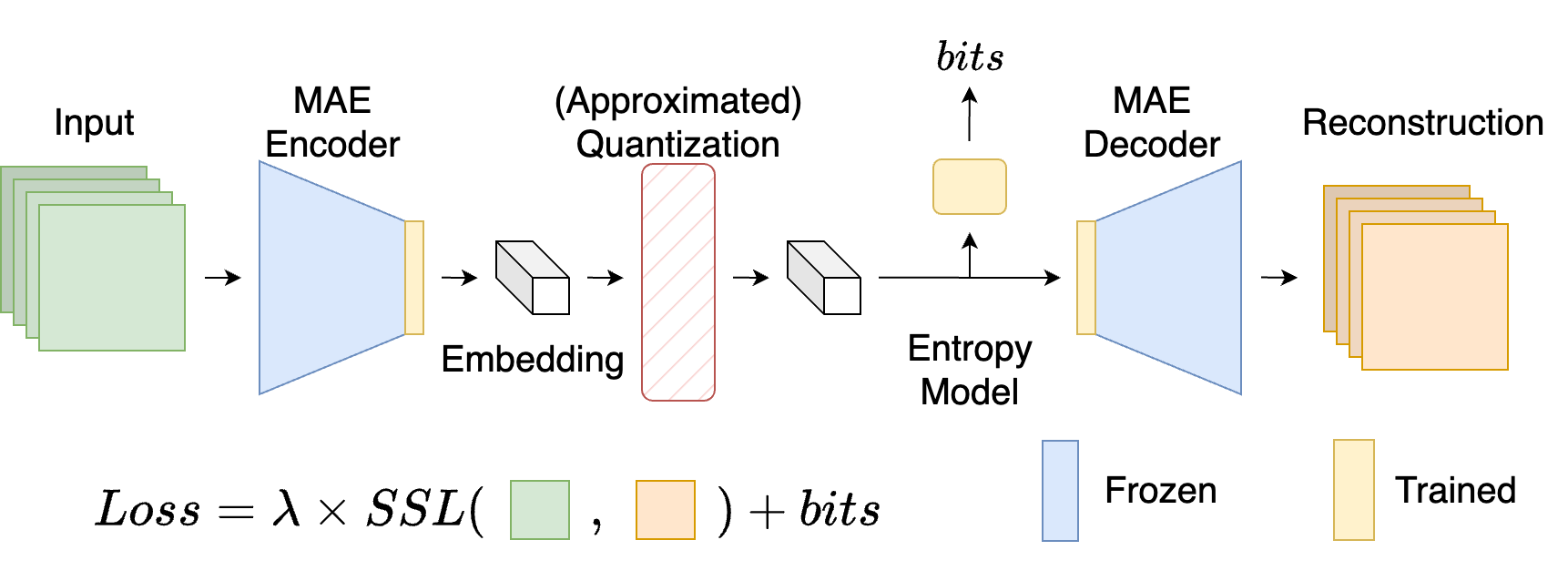}}
  \caption{Adapting an FM encoder for \textit{NEC}.}
  \label{fig:method}
\end{figure}

In this section, we frame our method as the optimization of a rate-distortion objective following the framework of end-to-end learned compression established in Ball\'{e} \emph{et al.}~\cite{balle2016end}.
Traditionally, these methods aim to transmit an input image or video using the smallest amount of bits possible (R - Rate) while minimizing the loss in quality (D - Distortion) after reconstruction by a decoder network.
This trade-off is controlled by $\lambda$, resulting in an optimization problem of the form $\min \lambda R + D$.

For \textit{NEC}, we are not specifically interested in reconstructing the original data. Instead, given an EO input, we aim to transmit the smallest amount of bits possible (R) so that we can perform model training or inference for various tasks, achieving the smallest loss (D) possible across them.
In what follows, we detail each component of our method, summarized in~\cref{fig:method}.

\textbf{Distortion Term --- SSL:}
Our distortion term is an aggregate loss over downstream tasks which are \textit{a priori} unknown.
Therefore, we require a proxy for this quantity.
In this work, we leverage self-supervised learning for this purpose, allowing for training on large unlabeled datasets to produce general-purpose embeddings which have shown competitive performance in many downstream tasks~\cite{ssl_remote_sensing}.

Specifically, we use MAE~\cite{mae} with a standard vision transformer~(ViT)~\cite{vit} architecture.
MAE masks part of the input and tasks the model with recovering it based on the unmasked input.
This is achieved by minimizing the mean squared error between the masked input and its reconstruction, which we take as our distortion term.
We choose MAE due to its simplicity and applicability to a large range of downstream tasks and the ViT-B architecture due to its wide use in recent EO models~\cite{prithvi, satmae, rvsa}.
However, any loss that encourages generally useful embeddings for EO and any compatible encoder architecture can be used.

\textbf{Rate Term:}
Ball\'{e}~\emph{et al.}~\cite{balle2016end} establishes the general methodology for computing the rate term in the full loss. We give a short overview and direct the reader to~\cite[Appendix 6.1]{balle2018variational} for more details.

Given an embedding vector $y$, the minimum number of bits required to transmit $y$ is given by Shannon's source coding theorem~\cite{shannon} as $-\log_2 p(y)$, where $p$ defines the probability distribution over all embedding vectors.
Our goal is to minimize the expectation of this quantity over the distribution of all embedding vectors $-\mathbb{E}\left[\log_2 p(y)\right]$, which is precisely the entropy of this distribution.
We adapt the fully factorized entropy model defined in Ball\'{e}~\emph{et al.}~\cite{balle2018variational} in order to model the probability distribution of these embeddings for MAE, leveraging the implementation in the CompressAI~\cite{begaint2020compressai} library. 

After patching, masking, and passing through the encoder, we obtain an embedding $y$ of shape $e \times n$, where $e$ is the embedding dimension of the encoder and $n$ depends on the input shape. 
We quantize $y$ by rounding to the nearest integer, using uniform noise as a differentiable proxy during training~\cite{balle2018variational}.
To model the distribution of $y$ we assume that all of its elements are independent and additionally that all elements with the same embedding dimension (the first dimension above) are identically distributed, resulting in $ p(y) = \prod_{e} \prod_{n} p_e(y_{e, n}) $.
%
Each $p_e$ is modeled using a small neural network.
%
%
%

\textbf{SSL Compression Loss:}
For an input $x$ for which the model produces the (approximated) quantized embedding $y$ and reconstruction $x\prime$, the loss is given as
\[ Loss(x, x\prime) = \lambda \cdot \text{MAE}(x, x\prime) - \log_2 p(y) \]
%
with $\lambda$ as the hyperparameter controlling the trade-off.

\textbf{Efficient Model Adaptation:} For training, we initialize our model with the pre-trained weights from Wang~\emph{et al.}~\cite{rvsa}, resulting from MAE training over the MillionAid~\cite{millionaid} dataset.
We freeze all layers of this model except for the patch embedding layers of the encoder and decoder, the final encoder layer, and the first decoder layer, optimizing only $\sim$10\% of the total parameters.

\label{sec:finetuning_setup}
\textbf{Fine-tuning}: For fine-tuning, the data consumer can train a model using the embeddings as surrogates for the raw training data.
The setup is illustrated in \cref{fig:downstream_finetuning}~(c).
We freeze the entire backbone, simulating the effect of the data consumer having access only to the embeddings.

Additionally, we experiment with the consumer having access to the first decoder layer (including the embedding layer).
In this case, we unfreeze these layers and apply them as the first step of processing the embeddings after entropy decoding and before the task-specific head.
The transfer of these layers' weights is a small one-time cost that would be paid \textit{a priori}, as with the probability models for the entropy codes~\cite{balle2016end}.

\section{EXPERIMENTS}
\label{sec:experiment}

For \textbf{Neural Embedding Compression (NEC)} we train on the MillionAid~\cite{millionaid} dataset for $20$ epochs, only $1.25\%$ of the 1600 epochs used for the original pre-training.
%
%
We use the same settings as Wang~\emph{et al.}~\cite{rvsa} except for a reduced learning rate of $1.5 \times 10^{-4}$ and 4 GPUs instead of 8, halving the total batch size to 1024.
Embeddings at different compression levels are obtained by training models with different values of $\lambda$ (ranging from $10^7$ to $10^{11}$) but with the same architecture. 
We benchmark \textit{NEC} against \textit{Raw Data Compression (RDC)} and \textit{Uniformly Quantized Embeddings (UQE)}, using these embeddings to train models for two downstream EO tasks: scene classification and semantic segmentation.

In \textbf{RDC}, we compare against transferring the raw data, after entropy coding using the popular Zstandard~\cite{zstd} algorithm, at 16 and 8-bit precision, and fine-tuning the entire model, as shown in \cref{fig:downstream_finetuning}~(a).
The 16-bit point is a useful reference despite the fact both curated datasets use 8-bit precision since many EO datasets (e.g., Sentinel 2) store data in greater than 8-bit precision.

In \textbf{UQE}, we compare against transferring quantized embeddings produced by the pre-trained ViT-B network from Wang~\emph{et al.}~\cite{rvsa}, as shown in \cref{fig:downstream_finetuning}~(b).
These are quantized using affine uniform quantization and entropy coded using Zstandard~\cite{zstd}. We show results for $32$ and $16$ bit float and $8$, $5$, $3$ and $2$ bit integer quantization.

Finally, we include \textbf{JPEG 2000}, a widely used and powerful image compression codec.
This approach is not entirely comparable to the others, as it aims to transmit the original image, rather than model-ready features, and thus still requires a large encoder to be employed for feature extraction at the consumer.
Nonetheless, it provides a good target for performance.
For our experiments, we use the original frozen model from Wang~\emph{et al.}~\cite{rvsa} as a feature extractor for these images.

\textbf{Downstream Task Evaluation:} For each task, we plot accuracy against the average size of an embedding for a sample from the validation set of the respective datasets.

\textit{1. Scene Classification:} We choose the UCMerced land use dataset~\cite{ucmerced} for our scene classification task.
It contains 21 classes, each with 100 images of size $256 \times 256$.
We use a simple architecture comprising a pooling layer followed by a single linear layer, discarding the [CLS] token.
For the pooling layer, we found the `AttentionPoolLatent' layer, as implemented in the `timm'~\cite{rw2019timm} library, to perform substantially better than other methods, such as average pooling.
We train all models with a learning rate of $1.25 \times 10^{-4}$ and a batch size of $32$ for 400 epochs using the AdamW~\cite{adamw} optimizer with weight decay of $0.05$.

\textit{2. Semantic Segmentation:} The Potsdam~\cite{potsdam} dataset is used for our semantic segmentation task.
This dataset contains large images of urban areas with semantic segmentation labels for six classes.
We process it as is standard~\cite{yamazaki2023aerialformer}, obtaining tiles of size $512 \times 512$. 
Unlike Wang~\emph{et al.}~\cite{rvsa}, we keep the `clutter' class and train with labels where the borders of semantic classes are ignored to avoid the ambiguity in labeling these.
Additionally, we restrict ourselves to the RGB channels to remain compatible with the MillionAid pre-training.
Most state-of-the-art models for semantic segmentation~\cite{rvsa,yamazaki2023aerialformer,cha2023billion} combine embeddings from different stages of the encoder.
While \textit{NEC} can be extended to compress multi-scale embeddings, we limit our investigation to the single embedding case.
We use a simple decoder composed of a first convolutional layer to reduce the embedding dimension by half, followed by six residual convolutional blocks.
Then, we use two PixelShuffle~\cite{pixelshuffle} blocks to upscale the result by a factor of four.
Finally, this output is passed on one branch to a main head with two convolutional layers with $128$ channels and on another branch to an auxiliary head with one convolutional layer with $64$ channels.

\section{Results}
%

\begin{figure}[t]
\centering
\begin{subfigure}[b]{\linewidth}
\includegraphics[width=\linewidth]{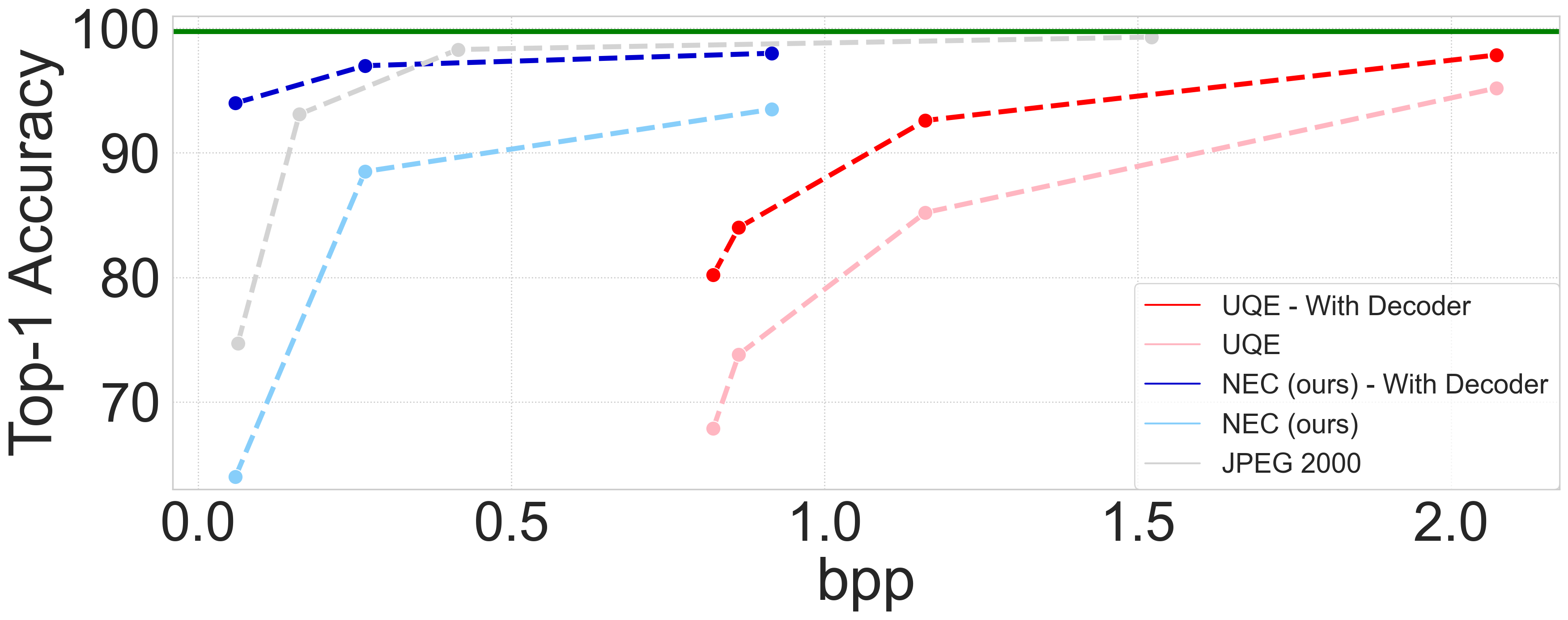}
  \caption{Classification accuracy on the UC Merced dataset}
  \label{fig:results_ucm}
\end{subfigure}
\centering
\begin{subfigure}[b]{\linewidth}
\includegraphics[width=\linewidth]{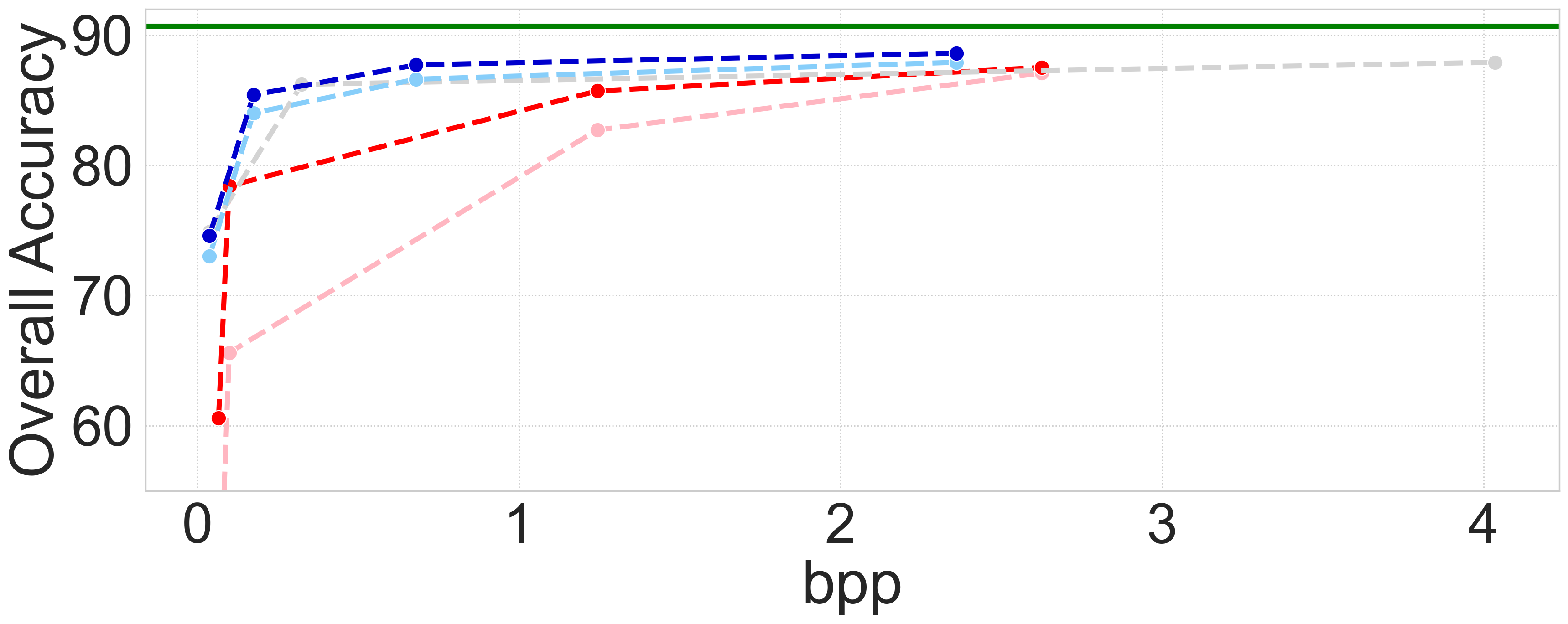}
  \caption{Overall accuracy for segmentation on the Potsdam dataset}
  \label{fig:results_potsdam}
\end{subfigure}
\caption{Accuracy metric vs bits per pixel. The green line shows uncompressed data and no frozen layers.}
\label{fig:results}
\end{figure}
\begin{figure}[t]
\centering
\includegraphics[width=0.6\linewidth]{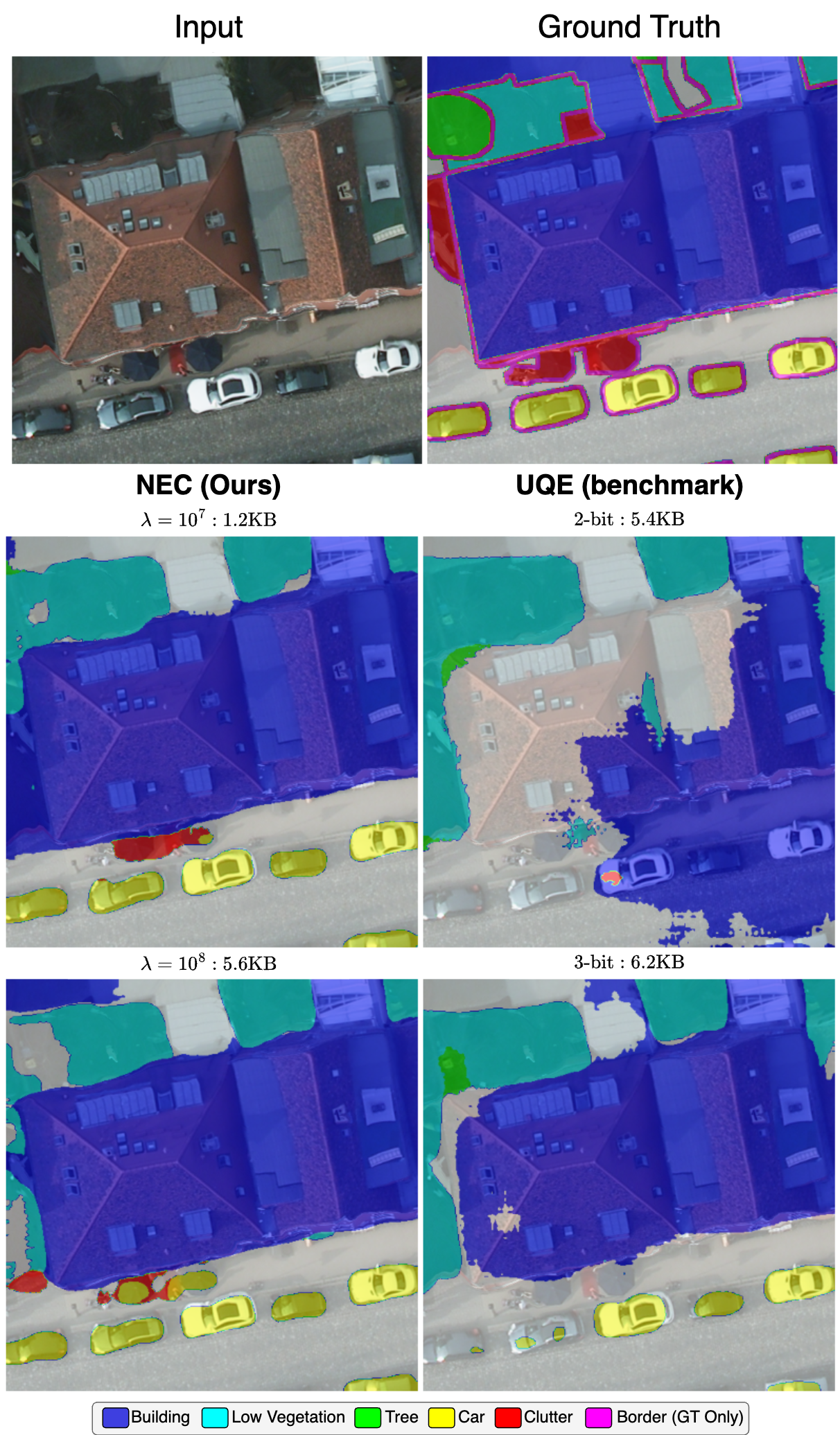}
  \caption{Semantic segmentation on the Potsdam dataset. The mean size of the embedding produced by the method over the validation set is shown in KB.}
  \label{fig:visual_results}
\end{figure}

\textit{1. Scene Classification:} \cref{fig:results_ucm} plots the results for the UCM Dataset. 
For the high-performance case, we observe that the simple \textit{UQE} strategy performs well until $5$ bit quantization (the rightmost point in the plot), however, for higher compression levels, the accuracy rapidly drops.
\textit{NEC} is able to achieve $94\%$ accuracy at an average embedding size of $0.47$KB, a reduction of only $5\%$ in accuracy for a size reduction of $\sim$$700$x from the raw data, greatly outperforming 3 and 2-bit \textit{UQE} as well as JPEG2000 in this bpp regime.
Retaining the first layer of the decoder is crucial to maintain performance at very high compression.
As the rate increases, our method delivers performance comparable to JPEG2000.

\textit{2. Semantic Segmentation:} Results are shown in \cref{fig:results_potsdam}.
\textit{NEC} scales better than \textit{UQE} with compression, achieving a $100$x reduction in size with only $5\%$ reduced accuracy, compared to a $12\%$ drop for \textit{UQE}.
\cref{fig:visual_results} shows a visual comparison of the inference outputs.
For a smaller bit budget on this difficult image, with many separate occurrences of classes to identify, \textit{NEC} produces more accurate segmentation masks.
In contrast to scene classification, our method only slightly outperforms JPEG 2000 through the whole bpp range, which may be explained by this task's greater reliance on high resolution image features.
However, JPEG 2000 requires a large encoder to be run as a feature extractor on the client side, whereas \textit{NEC} directly provides useful features.

\section{CONCLUSION}
\label{sec:conclusion}
  We introduce \textit{NEC}, a framework for downstream training and inference for EO tasks based on transmitting compressed embeddings.
  We show how any FM can be adapted to produce compressed embeddings with minimal compute.
  We achieve a data transfer reduction of one to three orders of magnitude compared to the original data, with only limited degradation in performance. 
  Additionally, we demonstrate performance comparable to or better than JPEG 2000.
  We argue the utility of our work from a sustainability point of view, enabling training and inference with limited bandwidth or computational resources as well as long-term storage of data from repositories, thus reducing data transfer, storage requirements, and their associated energy costs.
  Our codebase will be open-sourced at \url{https://github.com/IBM/neural-embedding-compression/}.

\clearpage
\bibliographystyle{IEEEbib}
\bibliography{refs}

\end{document}